%% file: root.tex
\documentclass[letterpaper, 10 pt, conference]{ieeeconf}  

\IEEEoverridecommandlockouts                              

\usepackage{amsmath} 
\usepackage{amssymb}  
\usepackage{bm}
\usepackage{units}
\usepackage{hyphsubst}
\usepackage{nicefrac}
\usepackage{paralist}
\DeclareMathOperator{\atantwo}{atan2}

\usepackage{array} 

\usepackage{graphics} 
\usepackage[pdftex]{graphicx}
\usepackage[pdftex]{xcolor}
\usepackage{subcaption}

\usepackage{tikz,pgfplots}
\usetikzlibrary{backgrounds,fit,calc,positioning,matrix,decorations.markings,arrows,arrows.meta,fadings,decorations.pathmorphing,shapes.geometric}
\usepgfplotslibrary{colormaps}

\input{defs.tex}

\title{\LARGE \bf
Vehicle Position Estimation with Aerial Imagery from\\Unmanned Aerial Vehicles
}

\author{Friedrich Kruber$^{1}$, Eduardo S\'{a}nchez Morales$^{1}$,\\Samarjit Chakraborty$^{2}$, Michael Botsch$^{1}$
\thanks{$^{1}$Technische Hochschule Ingolstadt, Research Center CARISSMA, Esplanade 10, 85049 Ingolstadt, Germany,
\{firstname.lastname\}@thi.de\
$^{2}$University of North Carolina at Chapel Hill (UNC), USA, \{firstname\}@cs.unc.edu}
}

\begin{document}

\maketitle
\thispagestyle{empty}
\pagestyle{empty}

\begin{abstract}
The availability of real-world data is a key element for novel developments in the fields of automotive and traffic research. Aerial imagery has the major advantage of recording multiple objects simultaneously and overcomes limitations such as occlusions. However, there are only few data sets available. This work describes a process to estimate a precise vehicle position from aerial imagery. A robust object detection is crucial for reliable results, hence the state-of-the-art deep neural network Mask-RCNN is applied for that purpose. Two training data sets are employed: The first one is optimized for detecting the test vehicle, while the second one consists of randomly selected images recorded on public roads. To reduce errors, several aspects are accounted for, such as the drone movement and the perspective projection from a photograph. The estimated position is comapared with a reference system installed in the test vehicle. It is shown, that a mean accuracy of \unit[20]{cm} can be achieved with flight altitudes up to \unit[100]{m}, Full-HD resolution and a frame-by-frame detection. A reliable position estimation is the basis for further data processing, such as obtaining additional vehicle state variables. The source code, training weights, labeled data and example videos are made publicly available. This supports researchers to create new traffic data sets with specific local conditions.
\end{abstract}

\section{INTRODUCTION}
\begin{figure}
	\vspace{2 mm}
		\centering
		\includegraphics[width=\columnwidth]{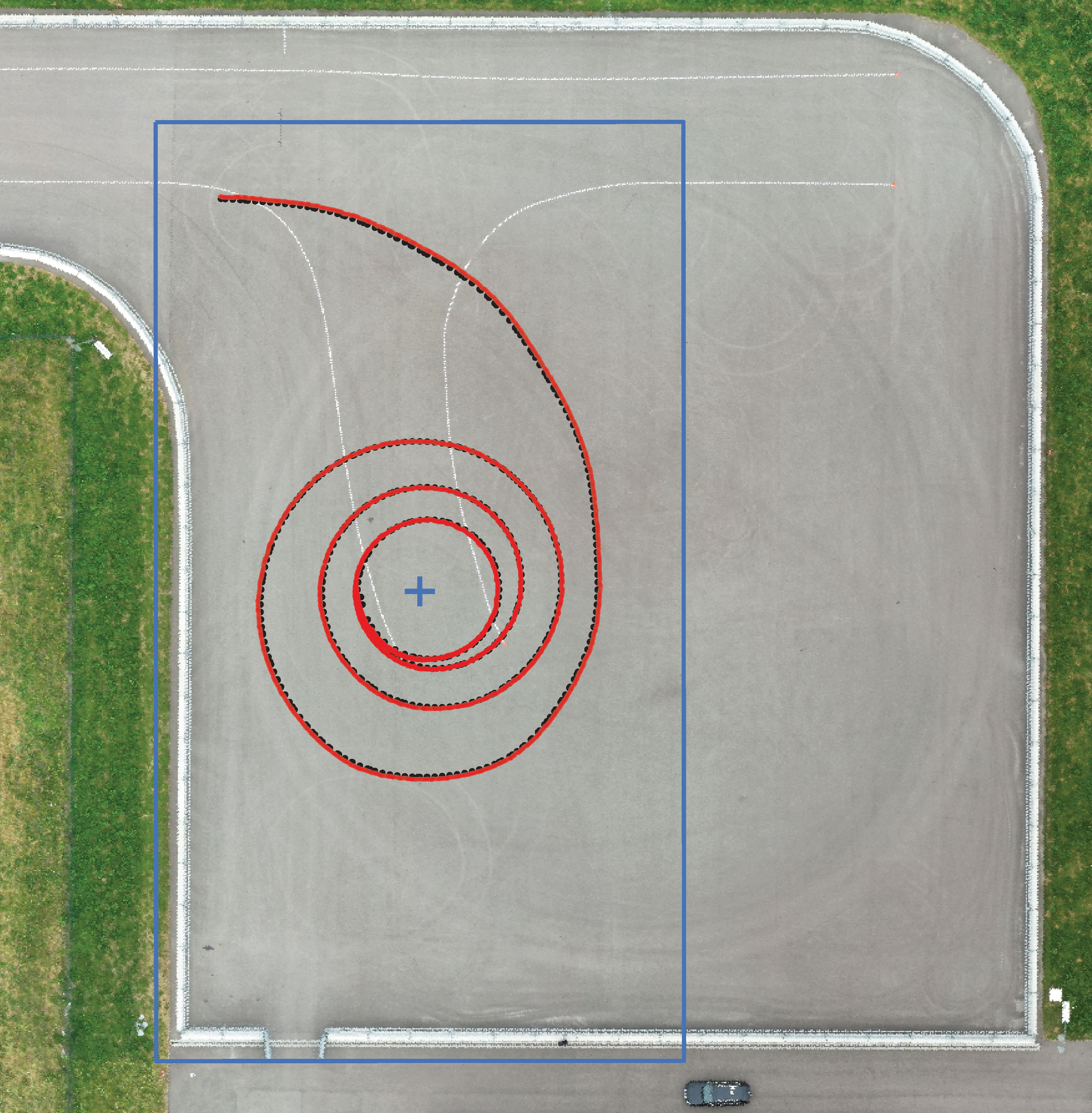}
		\caption{\small{Bird's eye view on an orthorectified map of the test track at \unit[50]{m} flight altitude. Depicted is the trajectory of the reference in black and drone in red, respectively. The car was driven in spirals to capture different vehicle poses and positions in the image frame. The blue rectangle and cross indicate the image frame and center. For estimation purposes, a true-to-scale T-junction is painted white on the test track.}} 
		\label{fig:example_trajectory}
		\vspace{-5 mm}
	\end{figure}

Real-world data is essential for automotive research and traffic analysis. The publicly available data sets can be mainly split in two groups: The first group provides data from the vehicle perspective as provided in the KITTI \cite{Geiger2013IJRR}, Waymo \cite{waymo_open_dataset} or Audi A2D2 \cite{aev2019} data sets. These kind of data sets boost research in terms of in-vehicle functionality, mainly computer vision tasks. In order to analyze traffic with an overall view on the situation, other approaches are preferable. Commonly used infrastructure sensing technologies like inductive loops provide accurate accumulated traffic data, while not being capable of providing individual trajectories of traffic participants. Aerial imagery from Unmanned Aerial Vehicles (UAV), usually drones, overcome this limitation. Yet, only few birds-eye-view data sets are available, while interest is growing due to technological progress. Recently published work and data sets are dicussed in Section \ref{relatedwork_section}. The behavior of traffic participants and infrastructural conditions differ throughout the world. This underlines the need for data according to the local specifics. 
The present work describes a process to generate reliable position data. Figure \ref{fig:example_trajectory} depicts an example from the experiments of Section \ref{results_section}. 

For further details on how to obtain additional vehicle state variables from UAV imagery, given the position estimation, the reader is referred to \cite{Morales20}.

Aerial remote sensing measurements have various advantages, such as dozens of objects can be captured in parallel with one sensor. UAVs are versatile by means of locations and covered area on ground. Also, the hovering position can be chosen to reduce oclusion. The generated data suits for both, research on individual traffic participant behavior and its predictions \cite{Morales2016,nadarajan2018machine}, as well as accumulated traffic flow analysis \cite{Neubert.2000, kruber2019highway}. 
Batteries are the bottleneck, but this disadvantage can nowadays be compensated by tethered systems, which allow flight durations of several hours. Wind and water resistance, alongside with low-light capabilities of cameras, are constantly improving, but remain a weak point.  
	
Before generating such birds-eye-view data, several aspects need to be taken into account: Drones are often equipped with non-metric cameras, so that the distortion has to be removed. The videos are affected by some movement and rotation of the hovering drone. Estimating the location of a vehicle within its environment requires a fixed frame. This property can be achieved algorithmically using image registration techniques. Photograhps yield a perspective projection, so that the detected objects are displaced compared to the ground truth. Section \ref{architecture_section} details how these aspects can be addressed.
\begin{figure*}[ht]
\vspace{2 mm}
\centering
	\input{ToolChainComplete.tex}
	\caption{\small{The overall process: From data recording to relief displacement correction.}}
	\label{fig:ToolChainComplete}
	\vspace{-0.1cm}
\end{figure*}
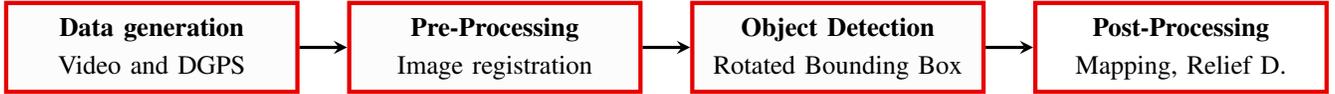
\subsection*{\bf{Own contributions}}
The three main contributions of the present work can be stated as follows: 

First, a framework to obtain precise vehicle positions from UAV imagery based on instance segmentation and image registration is provided. To the knowledge of the authors, no other comparable open sourced framework is available. 

Second, it is shown, how the accuracy can be optimized, compared to related work. Reducing the error is meaningful, for example, to associate a vehicle to its actual lane. Furthermore, a small error is necessary to detect lane changes at the right time instance and to compute criticality measures in a general sense. Accurate data acquisition is also essential to understand the locally characterized driving behavior and to develop algorithms based on it. A precise localization and representation of the vehicle's shape allows the use of simple trackers such as \cite{Bewley2016_sort}, which associates detections across frames in a video sequence. 

Third, the method's capabilities and limitations are evaluated with an industrial grade reference sytem. 
This work is a step towards large-scale colletion of traffic data using UAVs and discusses its feasibility and challenges.
\section{RELATED WORK}
\label{relatedwork_section}
Object detection and tracking via UAV gained attention over the past years. DroNet \cite{Kyrkou.2018} investigates the real time capability of vehicle detection with small onboard hardware. DroNet is a lean implementation of the YOLO network \cite{Redmon.2016}, where the number of filters in each layer is reduced. DroNet outputs several frames-per-second (fps) with onboard hardware, but at the cost of lower detection performance and image resolution. The network struggles with variations in flight height and vehicle sizes. It outputs horizontal bounding boxes, which are not suitable for estimating certain variables such as orientation.
The $R^3$ network \cite{R3.2018} enables the detection of rotated bounding boxes. $R^3$ is a bounding box detector, while Mask R-CNN \cite{he2017maskrcnn} yields instance segmentation. Reference \cite{Mou.2018} approaches vehicle detection via instance segmentation. One goal of \cite{Mou.2018} is to obtain a high detection rate at higher altitudes, while the present work pursues a precise position estimation at lower altitudes up to \unit[100]{m}. Also, the experimental results are compared to a reference system this paper. 

The methodology of \cite{GUIDO2016136} to assess aeriel remote sensing performance is comparable to this work. A test vehicle was equipped with a GPS logger to receive positions and speed. The images were geo-referenced to obtain a fixed frame. The main differences to the present work can be stated as follows:  The detection algorithm compares the differences between two frames, hence identifying moving objects by localizing altered pixel values. This type of detector is prone to errors, e.g., during vehicle standstills, changing light conditions or due to the movement of vegetation, as stated by the authors. The output is a non-rotated bounding box, which fails to estimate the vehicles shape and thereby worsens the position estimation. The reference sensor\footnote{Video VBOX Pro} in \cite{GUIDO2016136} provides accuracy of \unit[20]{cm} at best, assuming the Diffenrential GPS version. In the present work, the reference sensor's accuracy is \unit[1]{cm}, which is necessary to compare at pixel level, see Section \ref{data_recording}. The images in \cite{GUIDO2016136} were processed with a Gaussian blur filter, which is claimed to eliminate high frequency noise. Applying such a filter blurs the edges and is contraproductive when applying a neural network detector. Finally, relief displacement was not taken into account, which causes an increasing error with growing distance to the principal point, see Section \ref{relief_displacement}. The authors state a normalized root mean square error of \unit[0.55]{m} at a flight altitude of \unit[60]{m}. By the same measure, the error obtained in the present work is much lower with \unit[0.18]{m} at a flight altitude of \unit[75]{m} and identical image resolution. 

Except for DroNet, a missing publicly available implementation is common to all the above mentioned publications. 

Recently, the highD \cite{Krajewski.2018}, inD \cite{inDdataset} and INTERACTION \cite{interactiondataset} data sets were published. They provide processed traffic data obtained with drone and static camera images. While \cite{Krajewski.2018} provides German highway data, \cite{interactiondataset} offers urban sceneries like crossings and roundabouts and \cite{inDdataset} includes furthermore pedestrians and cyclists. The Stanford campus data set mainly captures pedestrians and bicycles on a campus, the publication focuses on human trajectory prediction \cite{Robicquet.2016}. While \cite{Krajewski.2018, inDdataset, interactiondataset} provide traffic scenario data sets, this paper describes the procedure to obtain vehicle positions and compares the results to a widely accepted reference. Finally, the code is open sourced for further improvements and to faciliate the generation of new data sets.
\section{METHOD}
\label{architecture_section}
Generating traffic data with UAVs is appealing, but certain challenges have to  be mastered. First of all, the images are recorded with a flying object, i.\,e. a fixed frame has to be established. Second, photographs yield a perspective projection. The top of objects are displaced from their bases in vertical recorded photgraphs, leading to a false interpretation, when directly computing positions from their bounding boxes. Obtaining bounding boxes in a sequence of many images, on the other hand, requires matured detection techniques, which are limited by the accuracy and amount of labeled training data. Finally, when performing a benchmark, the mapping and synchronisation of both data sources have to be considered.

In the following, the main steps are described as depicted in Figure \ref{fig:ToolChainComplete}. Beforehand, the coordinate systems used in this work are explained.

The vehicle moves on the Local Tangent Plane (LTP), where $x_{\text{L}}$ points east, $y_{\text{L}}$ north and $z_{\text{L}}$ upwards, with an arbitrary origin $o_{\text{L}}$ on the ground of the earth. The Local Car Plane (LCP) is defined according to the ISO\,8855 norm, where $x_{\text{C}}$ points to the hood, $y_{\text{C}}$ to the left seat, $z_{\text{C}}$ upwards, with the origin $o_{\text{C}}$ at the center of gravity of the vehicle. For simplification, it is assumed that 
\begin{inparaenum}[1)]
	\item the $x_{\text{C}}y_{\text{C}}$-plane is parallel to the $x_{\text{L}}y_{\text{L}}$-plane,
	\item the centre of mass is identical to the geometrical centre, and
	\item the sensor in the vehicle measures in the LCP.
\end{inparaenum}
The Pixel Coordinate Frame (PCF) is a vertical image projection of the LTP, where $x_{\text{P}}$ and $y_{\text{P}}$ represent the axes, with the origin $o_{\text{P}}$ in one corner of the image. Quantities expressed in PCF are given in pixels (px).

Throughout this work, vectors are represented in boldface and matrices in boldface, capital letters.  
\subsection{Data Recording}
\label{data_recording}
The data set was recorded on a test track. This gives degrees of freedom regarding arbitrary trajectories within the image frame. Experiments can be repeated with the same setup. On the other side, recording on public roads challenges the detector. Since test vehicles are a limited resource, a different approach is chosen to validate the detector to some extend. A second training set, recorded on public roads, is used to validate wheter the test vehicle is detected in a robust manner. See Table \ref{tb:AP_table} in Section \ref{results_section} for further details. The results suggest, that given a suitable large training data set, the detection on public roads also performs well. 

Next, the recording process is detailed.
Table \ref{tb:numFrames} depicts the flight altitudes and Ground Sampling Distance (GSD) for the drone\footnote{DJI Phantom 4 Pro V2} in use. 
The GSD is also known as photo scale or spatial resolution, see Section \ref{GCPs} for details about the computation. 
\begin{table}[h!]
\centering
\renewcommand{\arraystretch}{1.2}
\newcolumntype{?}{!{\vrule width 1pt}}
\begin{tabular}{r ? c c c}
	\hline 
  Flight altitude & \unit[50]{m} & \unit[75]{m} & \unit[100]{m} \\
  \hline
    Number of frames & 14\,532 & 15\,217 & 24\,106 \\
  \hline
  GSD [\unit{}{cm/px}] & 3.5 & 5.2 & 6.9 \\
  \hline 
\end{tabular}
\caption{\small{Total count of frames and GSD per altitude. At higher altitudes a larger area on ground has to be covered, thus increasing the number of video frames.}}
\label{tb:numFrames}
\vspace{-0.2 cm}
\end{table}

Generally, for a vertical photograph, the GSD $S$ is a function of the focal length $f$ of the camera and flight altitude $H$ above ground:
\begin{align}
S=\frac{f}{H}.
\label{eqn:ScaleFocalHeight}
\end{align}
Varying altitudes brings flexibility in the trade-off between GSD and captured area on the ground. The videos were recorded with \unit[50]{fps} and 4K resolution (\unit[3840]{px} x \unit[2160]{px}) and afterwards compressed to Full-HD (\unit[1920]{px} x \unit[1080]{px}) by applying a bicubic interpolation. For each altitude five videos were recorded.
The camera exposure time was kept constant at \unit[$\frac{1}{400}$]{s} for all recordings. 

A spiral template trajectory was driven to obtain different vehicle poses and to cover a large are of the image. The test vehicle was then equipped with a driving robot and a Satellite Navigation system\footnote{GeneSys ADMA-G-PRO+} that receives RTK correction data. This ensures an identical reproduction of the trajectory for all experiments, and a centimeter-accurate vehicle localization.  
\subsection{Pre-Processing}
\label{preprocess_section}
\begin{figure}
  \begin{minipage}[b]{0.24\textwidth}
    \includegraphics[width=\textwidth]{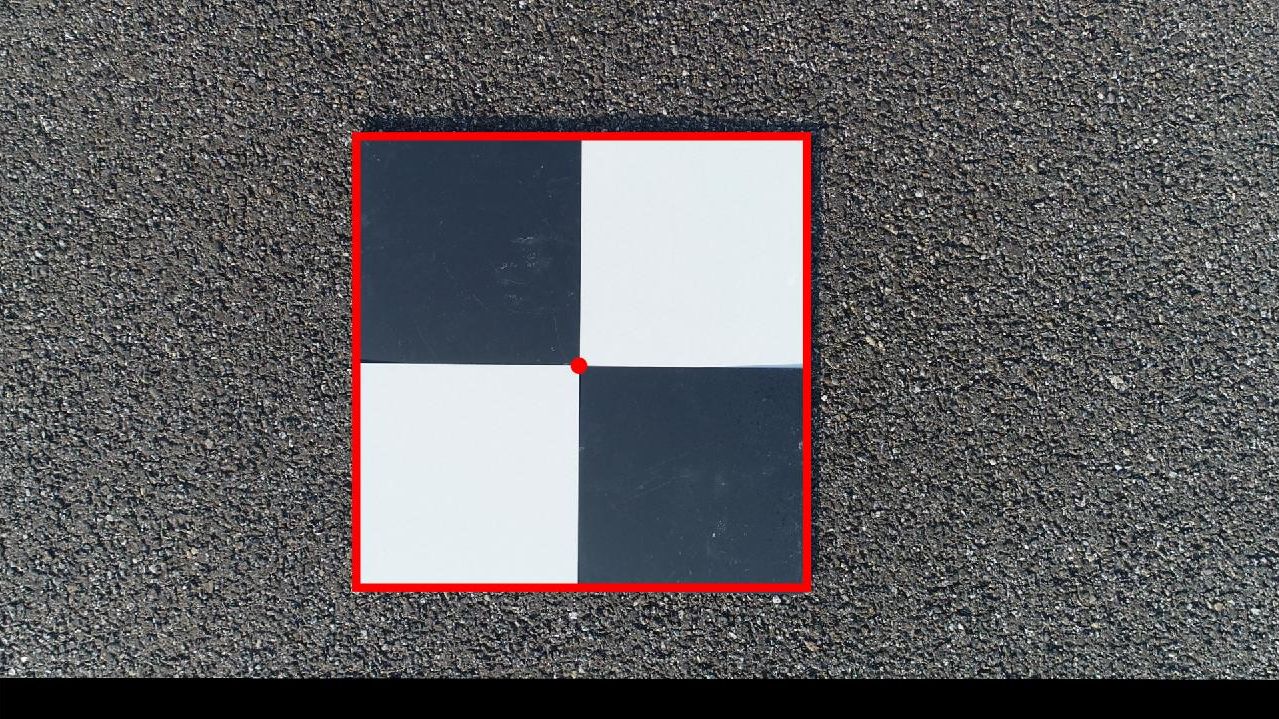}
  \end{minipage}
  \begin{minipage}[b]{0.24\textwidth}
    \includegraphics[width=\textwidth]{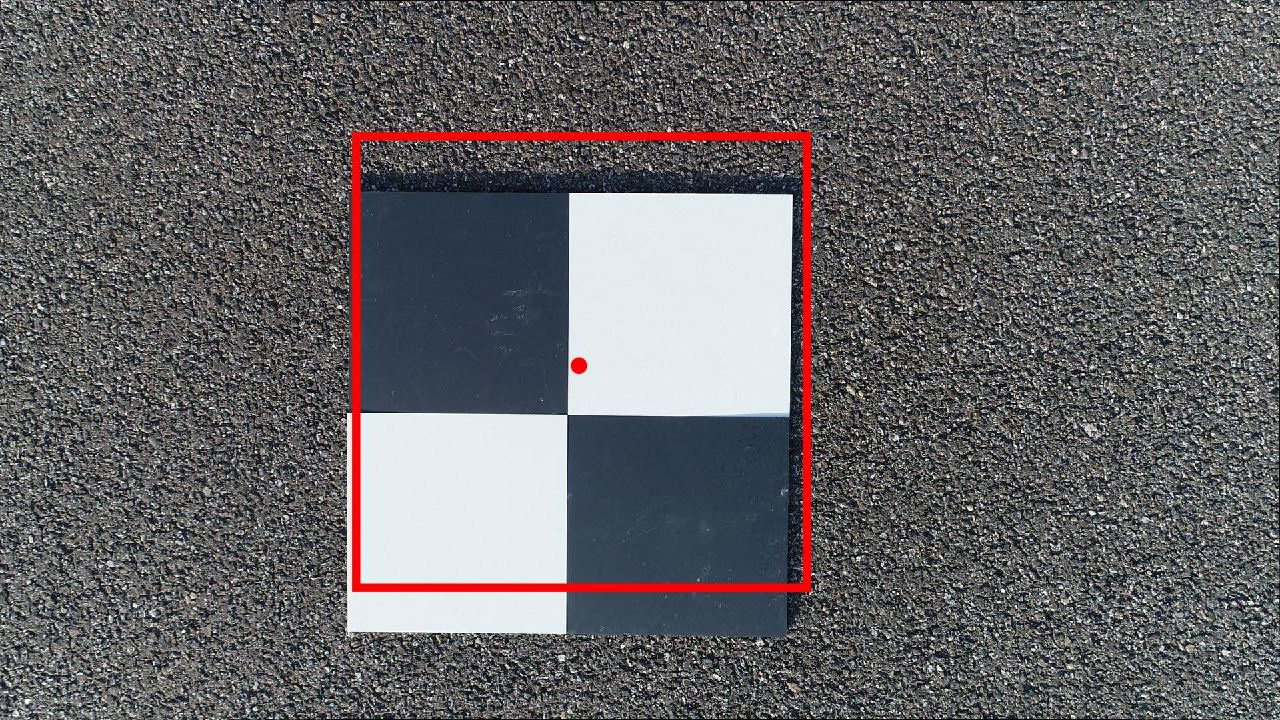}
  \end{minipage}
\caption{\small{Registered image (left) and the raw drone image (right) of a Ground Control Point (GCP) from \unit[1.5]{m} height. The left image borders are clipped due to translation and rotation. The red box depicts the object location from the first video frame, which was shot around \unit[30]{s} beforehand.}}
\vspace{-0.4cm}
\label{fig:registration}
\end{figure}

The pre-processing consists generally of two parts: The camera calibration and the image registration.
According to the manufacturer of the drone, the camera is shipped calibrated, so this step is skipped.
The image registration is performed to overlay the sequential frames over the first frame to ensure a fixed frame. The registration applied in this work is composed of a correction of the orientation, translation, and scaling of the image. Figure \ref{fig:registration} depicts an example of the registration result. 
This process involves three steps in order to find correspondences between two images: a feature detector, a descriptor and finally the feature matching. The goal of the detector is to find identical points of interest under varying viewing conditions. The descriptor is a feature vector, which describes the local area around the point of interest. For this work, the Speeded Up Robust Features algorithm \cite{Bay.2006} is used as a detector and descriptor. To match the points between two images, the distances between the feature vectors are computed. If the distance fulfills a certain criterion, e.\,g., a nearest neighbor ratio matching strategy, a matching point on two images was found. 
The matches are then fed into the MLESAC algorithm \cite{MLESAC.2000} to eliminate outliers. Lastly, a randomly selected subset of the remaining matching points are used for the image scaling, rotation and translation. 
\subsection{Object detection}
\label{detection_section}
The pre-processed images are fed into the Mask-RCNN implementation of \cite{matterport_maskrcnn_2017}, which is pre-trained on the Common Objects in Context (COCO) data set \cite{coco.2014}. Mask R-CNN extends Faster R-CNN \cite{NIPS2015_5638} by adding a parallel, Fully Convolutional Network \cite{FullyConvNetwork} branch for instance segmentation, next to the classification and bounding box regression from Faster R-CNN. The network is a so called two stage detector: In the first stage, feature maps generated by a backbone network are fed into a Region Proposal Network, which outputs Regions of Interest (RoI). In the second stage, the predicitions are performed within the RoIs. Additionally, a Feature Pyramid Network is included for detecting objects at different scales \cite{FeatPyramidNetwork}. 

Transfer learning has been applied with two training sets: The first one contains 196 randomly selected and manually labeled images with the test vehicle being present on all images. The second set contains 133 randomly selected images with 1\,987 manually labeled vehicles, recorded on public roads. This is done to examine the performance under general conditions. In the following, the two training sets are named "specialized" and "general". Details about the training procedure can be obtained from the code.
Finally, the binary mask output is used to compute the smallest rectangle containing all mask pixels using an OpenCV \cite{opencv_library} library.
\subsection{Post-Processing steps}
\label{postprocess_section}
To complete the process, two more steps are performed. First, the output from the detector, given in PCF, has to be mapped on the LTP. Finally, the disturbing relief displacement is handled.

\subsubsection{PCF Mapping}
\label{GCPs}
A comparison to the reference requires the mapping of the PCF on the LTP. For this, GCPs are placed on the $x_{\text{L}}y_{\text{L}}$-plane, in such a way that they are visible on the PCF. The \textit{i}-th GCP is defined in LTP as
$\boldsymbol{g}_{i,\text{L}}=
\begin{bmatrix}
x_{i,\text{L}}&
y_{i,\text{L}}
\end{bmatrix}^\text{T}$,
and in PCF as
$\boldsymbol{g}_{i,\text{P}}=
\begin{bmatrix}
x_{i,\text{P}}&
y_{i,\text{P}}
\end{bmatrix}^\text{T}$.
The GSD $S$ is calculated from two GCPs by
\begin{align}
S=\frac{\left|{\boldsymbol{g}_{i+1,\text{L}}-\boldsymbol{g}_{i,\text{L}}}\right|}{\left|{\boldsymbol{g}_{i+1,\text{P}}-\boldsymbol{g}_{i,\text{P}}}\right|}.
\label{eqn:ScalePCF}
\end{align}

The \textit{i}-th GCP can then be expressed in meters by
$\boldsymbol{\tilde{g}}_{i}=\boldsymbol{g}_{i,\text{P}}\cdot S=
\begin{bmatrix}
\tilde{x}_{i}&
\tilde{y}_{i}
\end{bmatrix}^\text{T}$.
The orientation offset $\delta$ from the LTP to the PCF is calculated as
$\delta=\theta_{i}-\theta_{i,\text{L}}$, with
$\theta_{i}=\atantwo(\tilde{y}_{i+1}-\tilde{y}_{i},\: \tilde{x}_{i+1}-\tilde{x}_{i})$. 
$\theta_{i,\text{L}}$ is calculated  by analogy.
The GCP $\boldsymbol{\tilde{g}}_{i}$ is then rotated as follows
\begin{align}
\boldsymbol{\hat{g}}_{i}=\boldsymbol{R}\left(\delta\right)^\text{T}\boldsymbol{\tilde{g}}_{i},
\label{eqn:GCPorientedinLTP4}
\end{align}
where $\boldsymbol{R}\left(\cdot\right)$ is a 2D rotation matrix.
The linear offsets from the LTP to the PCF are calculated by
$\boldsymbol{\Delta}=\boldsymbol{\hat{g}}_{i}-\boldsymbol{g}_{i,\text{L}}$. 
Finally, a pixel $\boldsymbol{p}_{\text{P}}=\begin{bmatrix}
x_{\text{P}}& 
y_{\text{P}}
\end{bmatrix}^\text{T}$ on the PCF can be mapped to the LTP by
\begin{equation}
\boldsymbol{p}_{\text{P}}^{\text{L}}=\left(\boldsymbol{R}\left(\delta\right)^\text{T}\left(\boldsymbol{p}_{\text{P}}\cdot S \right)\right)-\boldsymbol{\Delta}\text{.}
\label{eqn:GCPorientedinLTP6}
\end{equation}

The next stage is to semantically define the four bounding box corners.
It is assumed that the box covers the complete shape of the vehicle.
The \textit{i}-th corner of the bounding box is defined in PCF as 
$\boldsymbol{b}_i=
\begin{bmatrix}
x_{\text{b},i,\text{P}}&
y_{\text{b},i,\text{P}}
\end{bmatrix}^\text{T}$,
and the bounding box is defined in PCF as 
\begin{align}
\boldsymbol{B}_{\text{P}}=
\begin{bmatrix}
\boldsymbol{b}_1&
\boldsymbol{b}_2&
\boldsymbol{b}_3&
\boldsymbol{b}_4
\end{bmatrix}.
\label{eqn:BoundingBox2}
\end{align}
The corners of the bounding box are mapped to the LTP as shown in Eq. (\ref{eqn:GCPorientedinLTP6}) to obtain $\boldsymbol{B}_{\text{P}}^{\text{L}}$. 
The geometric centre of the vehicle $\boldsymbol{o}_{\text{veh}}$ is calculated by


\begin{align}
\boldsymbol{o}_{\text{veh}}=
\begin{bmatrix}
\frac{\max\left(\boldsymbol{B}_{\text{P}\,1,i}^{\text{L}}\right)+\min\left(\boldsymbol{B}_{\text{P}\,1,i}^{\text{L}}\right)}{2}\\
\frac{\max\left(\boldsymbol{B}_{\text{P}\,2,i}^{\text{L}}\right)+\min\left(\boldsymbol{B}_{\text{P}\,2,i}^{\text{L}}\right)}{2}
\end{bmatrix}\text{,}
\label{eqn:geomcentercalc}
\end{align}
for $i = 1,\dots,4$.
The dimensions of the detected vehicle are calculated next. 
Let
\begin{equation}
||{\boldsymbol{b}_2}-\boldsymbol{b}_1||<||{\boldsymbol{b}_3}-\boldsymbol{b}_1||<||{\boldsymbol{b}_4}-\boldsymbol{b}_1||\text{,}
\label{eqn:geomcentercalc2b}
\end{equation}
then $\hat{w} = S \cdot ||{\boldsymbol{b}_2}-\boldsymbol{b}_1||$ and $\hat{l} = S \cdot ||{\boldsymbol{b}_3}-\boldsymbol{b}_1||$ are the estimated width $\hat{w}$ and length $\hat{l}$ of the vehicle in meters.

		
\subsubsection{Relief displacement}
\label{relief_displacement}
\begin{figure}
\vspace{2 mm}
\centering
  \begin{minipage}[b]{0.45\textwidth}
    \includegraphics[width=\textwidth]{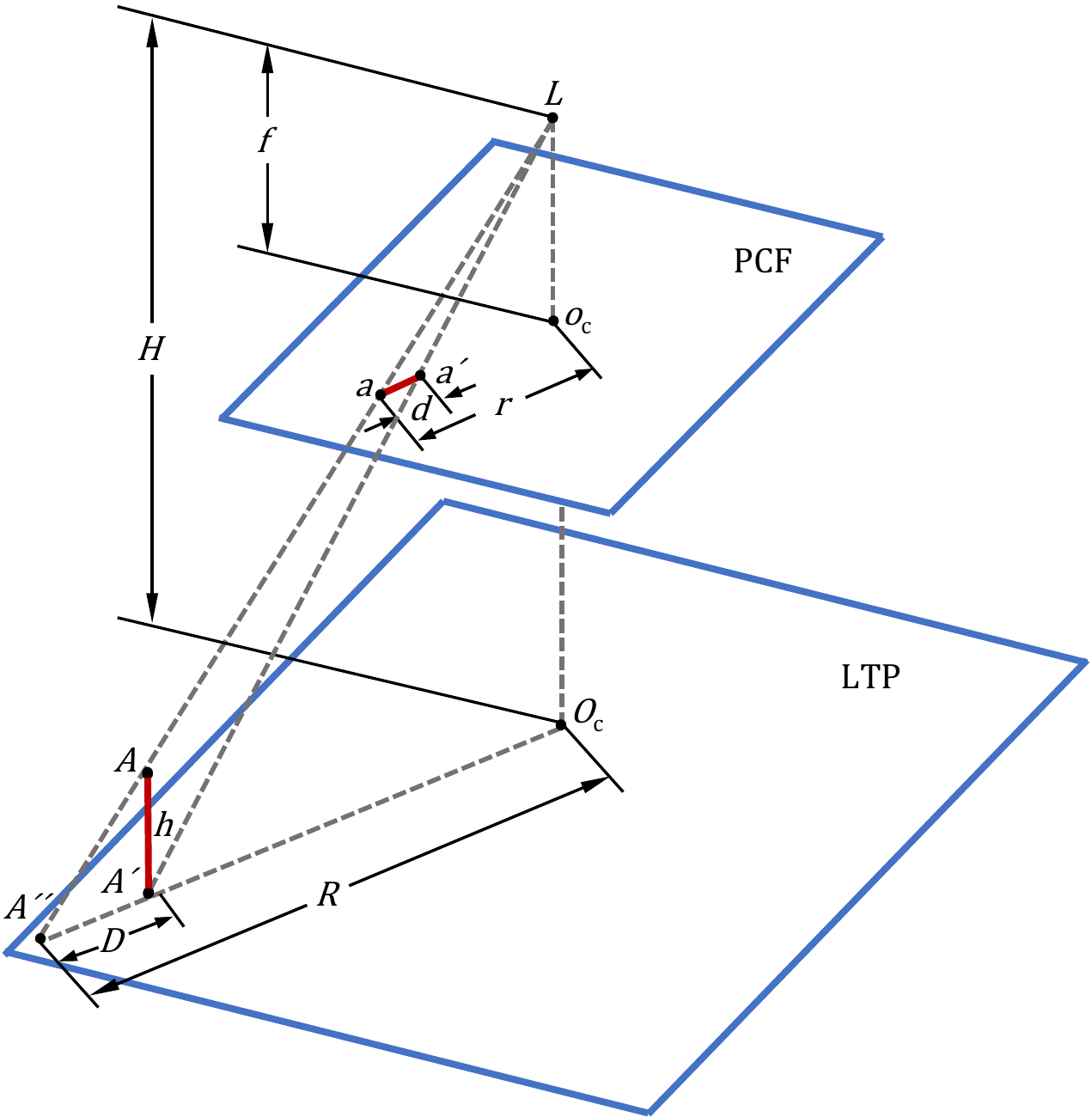}
  \end{minipage}
\caption{\small{Geometry of the relief displacement, adapted from \cite{Lillesand2003}. The red bar depicts an object of height $h$. Due to the perspective projection and $ R > 0$, the top of the bar $A$ is displaced on the photo compared to the bottom $A'$. The relief displacement $d$ is the distance between the corresponding points $a$ and $a'$ in the PCF.}}
\label{fig:Relief_distortion}
\vspace{-0.2cm}
\end{figure}
Photographs yield a perspective projection. A variation in the elevation of an object results in a different scale and a displacement of the object. An incrase in the elevation of an object causes the position of the feature to be displaced radially outwards from the principal point $O_{\text{c}}$ \cite{Lillesand2003}.

Assuming a vertical camera angle, the displacement can be computed from the similar triangles $LO_{\text{c}}A''$ and $AA'A''$, according to Figure \ref{fig:Relief_distortion}:
	\begin{align}
\frac{D}{h} = \frac{R}{H}\text{,} \quad \frac{d}{h} = \frac{r}{H}\text{,}
	\label{eqn:relief_displacement}
	\end{align}
where the second equation is expressed in GSD, with $d$ defining the relief displacement and $r$ the radial distance between $o_{\text{c}}$ and the displaced point $a$ in PCF. $D$ defining the equivalent distance of $d$, projected on ground, $R$ the radial distance from $O_{\text{c}}$, $H$ flight altitude and $h$ being the object height in LPT. $L$ is the camera lense exposure station, where light rays from the object intersect before being imaged at the cameras' sensor. The relief displacement decreases with an increasing hovering altitude and is zero at $O_{\text{c}}$.	

According to Eq. \ref{eqn:relief_displacement}, the bounding box has to be shifted radially. Two approaches are described: The first one requires knowledge of the vehicle sizes, and the second one is an approximation for unknown vehicle dimensions.
Since the training is performed to detect the complete vehicle body, the corner closest to $O_{\text{c}}$ can be usually identified as the bottom of the vehicle. So the height of this point is equal to the ground clearance. Knowing the height of this corner, its displacement is corrected as described in the following. 

Defining the horizontal and vertical resolution of the image as $r_{\text{x}}$ and $r_{\text{y}}$, the coordinates in PCF of $\boldsymbol{b}_i$ w.r.t. the image center are given by
\begin{equation}
\begin{bmatrix}
x_{\text{b},i,\text{img}}\\
y_{\text{b},i,\text{img}}
\end{bmatrix}=
\begin{bmatrix}
x_{\text{b},i,\text{P}}-\frac{r_{\text{x}}}{2}\\
y_{\text{b},i,\text{P}}-\frac{r_{\text{y}}}{2}
\end{bmatrix}\text{.}
\label{eqn:perspective3}
\end{equation}
The shift $\Delta_{x,\text{P}}$ along the $x_{\text{P}}$ axis is calculated on the PCF by	
\begin{equation}
\Delta_{x,\text{P}}=\frac{x_{\text{b},i,\text{img}}\cdot h_{\text{b},i,\text{L}}}{H}\text{,}
\label{eqn:perspective2}
\end{equation}
where $h_{\text{b},i,\text{L}}$ is the height of the \textit{i-th} corner on the LTP. The shift for $\Delta_{y,\text{P}}$ is computed by analogy along the $y_{\text{P}}$ axis. The approximated coordinates $\boldsymbol{b}_{i,\text{shift}}$ of $\boldsymbol{b}_i$ are then given by
\begin{equation}
\boldsymbol{b}_{i,\text{shift}}=\boldsymbol{b}_i-\begin{bmatrix}
\Delta_{x,\text{P}}&
\Delta_{y,\text{P}}
\end{bmatrix}^\text{T}\text{.}
\label{eqn:perspective4}
\end{equation}
Let $w$ be the width and $l$ the known length of the vehicle and $\boldsymbol{b}_1$ be the closest corner to the image centre. Then, $\boldsymbol{b}_1$ is used for scaling $\boldsymbol{b}_\text{l}$ and $\boldsymbol{b}_\text{w}$ as follows
\begin{equation}
\boldsymbol{b}_{\text{w},\text{scaled}}=\left(\frac{w}{\hat{w}}\cdot\left(\boldsymbol{b}_\text{w}-\boldsymbol{b}_1\right)\right)+\boldsymbol{b}_1\text{, and}
\label{eqn:boxbuild1}
\end{equation}
\begin{equation}
\boldsymbol{b}_{\text{l},\text{scaled}}=\left(\frac{l}{\hat{l}}\cdot\left(\boldsymbol{b}_\text{l}-\boldsymbol{b}_1\right)\right)+\boldsymbol{b}_1\text{,}
\label{eqn:boxbuild2}
\end{equation}
where $\text{w}$ is the element of $\boldsymbol{B}_{\text{P}}$ associated with $||{\boldsymbol{b}_2}-\boldsymbol{b}_1||$ and $\text{l}$ associated with $||{\boldsymbol{b}_3}-\boldsymbol{b}_1||$, respectively. The shifted centre of the vehicle can then be calculated by
\begin{equation}
\boldsymbol{o}_{\text{veh,shift}}=\frac{\boldsymbol{b}_{\text{w,}\text{scaled}}+\boldsymbol{b}_{\text{l,}\text{scaled}}}{2}\text{.}
\label{eqn:boxbuild3}
\end{equation}  
When gathering data on public roads, the vehicle dimensions are unknown and can not be estimated with a mono camera. An approximation  for the displacement can be performed by assuming that two corners of the bounding box, which form $\hat{l}$ and one of the corners is the closest to $o_{c},$ the ground clearance is usually visible. The height of the ground clearance can be approximated with \unit[15]{cm} for passenger cars \cite{TUVRideHeight}. The remaining two corners can usually be referred to as the vehicle body shoulders, which protrude further than the roof of the vehicle. The shoulders height is roughly half of the vehicle height and can be approximated with \unit[75]{cm} for passenger cars. Then all four corners can be shifted following Eq. (\ref{eqn:perspective2}). Although this is only a coarse approximation, the overall error is reduced compared to the initial situation of neglecting the displacement.
\section{EXPERIMENTS}
\label{results_section}
In this section the main potential sources of errors are being discussed. Later, the results of estimating the position will be presented. 
\subsection{Sources of errors}
\label{sourcesoferrors}
The overall process involves several steps, of which all affect the accuracy of the position estimation.
Two groups of potential errors can be distinguished: The first group describes general issues appearing from aerial imagery captured by an UAV. Here, the registration and relief displacement have the most significant influence. The second group is only of concern, when comparing to a reference system. In this group, maximizing the distance between the GCPs and localizing them precisely on the PCF is essential.
The following enumeration lists the main potential sources of errors:
\begin{itemize}
\item Camera calibration and image registration,
\item Image compression,
\item Camera exposure time,
\item Training data generation and object shape detection,
\item Location of the shape boundary on the vehicle body,
\item Rotation and GSD with GCPs,
\item Sensor synchronisation.	
\end{itemize}
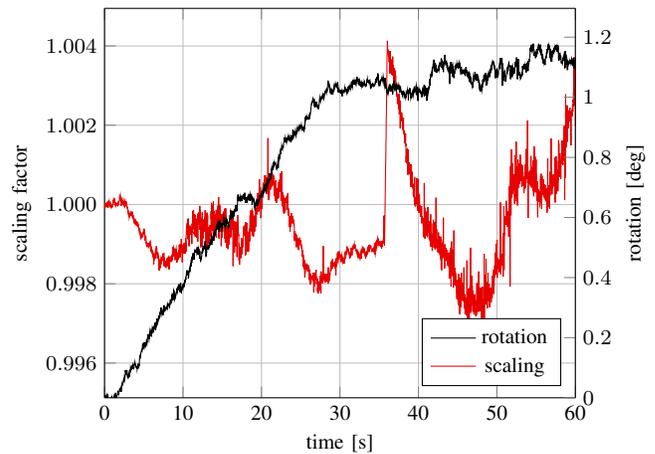
\begin{figure}
\vspace{2 mm}
  \begin{minipage}{0.45\textwidth}
	\input{scale_and_theta_registration.tex}
  \end{minipage}
  	\vspace{-0.6cm}
\caption{\small{Image registration: Scaling and rotation for a period of \unit[60]{s}: Rotation in \ref{pl:theta_reg} till \unit[30]{s}. The scaling factor in \ref{pl:scale_reg} with a drop at around \unit[35]{s}, caused by an altitude drop.}}
\label{fig:scale_and_theta_registration}
\vspace{-0.4cm}
\end{figure}
Wide angle lenses have the preferable focal length to capture a large area on ground. They are usually affected by barrel distortion, which decreases the GSD with increasing distance from the optical axis $O_{\text{c}}$. According to the drone manufacturer, the camera in use performs the corrections automatically. 
Every pixel deviation in the feature detection and matching during the image registration process inevitably leads to a deviation in rotation and GSD. Changing light conditions and the slight movements of the hovering drone affect the perception and thus influence the matching. Figure \ref{fig:scale_and_theta_registration} depicts a typical example of rotation and scaling, recorded over a period of one minute at \unit[100]{m} altitude. While achieving robust results, some potential outliers with a magnitude of approximately \unit[0.1]{\%} can be observed for the scaling parameter, which translates into a deviation of up to \unit[12]{cm} at \unit[100]{m} altitude. Filtering these variables was omitted to examine the robustness of the image registration algorithms.

The images are compressed in two ways. First, the resolution is reduced by half for both axes. Second, storing the images as JPEGs leads to lossy image compression. For example, smooth transitions can be found, which reduce the sharpness of edges. 
Another component is the camera exposure time. With exposure time, motion blur is induced, which can stretch the vehicle on the image or blurs the edges. Therefore, short exposure times are preferable, but at the cost of less light exposure. 

A robust semantic segmentation is crucial for achieving reliable results. Even though Mask-RCNN provides excellent results, see Section \ref{detection_performance_section}, minor deviations of at least one pixel can not be avoided. The deviations stem mainly from the manual image labeling and limited training data.

With an radially increasing distance from $O_{\text{c}}$, the relief displacement has a major influence on the results. The problem in correcting the displacement of vehicles is rooted in the complex shapes and varying heights. It is difficult to determine which part of the car has been detected exactly, even when inspected by a human: Assuming a flight altitude of \unit[100]{m} and a vehicle detected close to the image border, e.\,g., a distance of \unit[60]{m} to $O_{\text{c}}$, the displacement increases with \unit[0.6]{cm} per \unit{}{cm} change in object height. Detecting a feature of the vehicle at \unit[30]{cm} height, instead of the body bottom (Section \ref{relief_displacement}), yields an error of \unit[9]{cm}. 
 
The last two sources of error are only of concern when comparing the results with a reference system. 
The mapping of the PCF to the LTP is based on localizing the GCPs, see Section \ref{architecture_section}. Due to the limited resolution and image compression, a mis-locatization of typically one pixel per GCP in the PCF can be induced. Hence, all variables concerning the mapping, namely the GSD $S$, orientation offset $\delta$ and linear offset $\boldsymbol{\Delta}$, see Section \ref{GCPs}, are affected. The resulting error depends on the distance between two GCPs, hence $|\boldsymbol{g}_{i+1,\text{L}}-\boldsymbol{g}_{i,\text{L}}|$ should be maximized. 

Synchronisation between the two sensors is attained via UTC time stamps. Since UTC time stamps can not be associated to a certain frame for the drone in use, a LED signal, triggered by the Pulse-per-second signal (PPS) from a satellite navigation receiver, was recorded. This appears to be the best solution, since the circuit delay within the receiver and the LED rising time can be neglected. The first video frame showing the illuminated LED is associated with the corresponding UTC time stamp. Hence, the maximum synchronisation error is limited here to $\frac{1}{\text{fps}} = \unit[20]{\text{ms}}$. During the experiments, the maximum speed was around \unit[30]{km/h} leading to a worst case error of \unit[17]{cm} due to synchronisation.
\subsection{Detection performance}
\label{detection_performance_section}
A set of 50 images has been labeled for evaluation. Table \ref{tb:AP_table} depicts the Average Precision (AP) results according to the PASCAL VOC ($\mathrm{AP@IoU}=0.5$) \cite{Everingham10} and COCO ($\mathrm{AP@IoU}$[0.5:0.05:0.95]) \cite{coco.2014} definitions, where the Intersection over Union threshold is abbreviated as IoU. 
\begin{table}
\vspace{2 mm}
\centering
\renewcommand{\arraystretch}{1.2}
\newcolumntype{?}{!{\vrule width 1pt}}
\begin{tabular}{r ? c c ? c c}
  \hline
  Weights &  \multicolumn{2}{c}{Specialized} \vline \vline & \multicolumn{2}{c}{General} \\
  \hline
  Metric & \scriptsize{$\mathrm{AP}@0.5$} & \scriptsize{$\mathrm{AP}@[0.5,0.95]$} & \scriptsize{$\mathrm{AP}@0.5$} & \scriptsize{$\mathrm{AP}@[0.5,0.95]$}\\
  \hline
  \hline 
  \unit[50]{m} & 1.00 & 0.89 & 1.00 & 0.84\\
  \hline
  \unit[75]{m}  & 1.00 & 0.89 & 1.00 & 0.82 \\
    \hline
  \unit[100]{m} & 1.00 & 0.91 & 1.00 & 0.86\\
    \hline
\end{tabular}
\caption{\small{AP evaluation for all altitudes and both training sets.}}
\vspace{-0.4cm}
\label{tb:AP_table}
\end{table}
The detection is robust and the AP is similar for all three flight heights. An $\mathrm{AP@IoU}=0.5$ of 1 exhibits a detection rate of \unit[100]{\%} for the evaluation images. That holds for all images (see. Table \ref{tb:numFrames}) detected with the specialized training weights. For the general weights, the detection rate is \unit[99.97]{\%} w.r.t to the images listed in Table \ref{tb:numFrames}. This is reasonable, since the environment of the test track does not exhibit structures to be confused with a top-down view of a vehicle shape.  
It should be noted, that the detection performance does not directly reflect the accuracy of the position estimation, since the bounding box is computed according to the most outer pixels of the shape. Assuming the most outer pixels are detected and they do match to the actual vehicle body border, the position could still be computed correctly on a pixel level, although the IoU is less than one.
\subsection{Position estimation}
\begin{table*}[h!]
\vspace{2 mm}
\centering
\renewcommand{\arraystretch}{1.2}
\newcolumntype{?}{!{\vrule width 1pt}}
\begin{tabular}{r ? c  c  c  c ? c  c  c  c ? c  c  c  c}
  \hline
  Altitude & \multicolumn{4}{c}{\unit[100]{m}} \vline \vline & \multicolumn{4}{c}{\unit[75]{m}} \vline \vline & \multicolumn{4}{c}{\unit[50]{m}} \\
  \hline
  Weights & \multicolumn{2}{c}{\footnotesize{Specialized}} & \multicolumn{2}{c}{\footnotesize{General}} \vline \vline & \multicolumn{2}{c}{\footnotesize{Specialized}} & \multicolumn{2}{c}{\footnotesize{General}} \vline \vline & \multicolumn{2}{c}{\footnotesize{Specialized}} & \multicolumn{2}{c}{\footnotesize{General}}\\
  \hline 
  Corrections & \footnotesize{reg} & \footnotesize{reg+shift} & \footnotesize{reg} & \footnotesize{reg+shift} & \footnotesize{reg} & \footnotesize{reg+shift} & \footnotesize{reg} & \footnotesize{reg+shift} & \footnotesize{reg} & \footnotesize{reg+shift} & \footnotesize{reg} & \footnotesize{reg+shift} \\
  \hline
  \hline
Median [px] & 3.27 & 2.41 & 3.33 & 2.71 & 3.26 & 2.70 & 3.67 & 3.08 & 4.47 & 3.93 & 4.14 & 4.05 \\
  \hline
Mean [px] & 3.87 & 2.95 & 3.96 & 3.27 & 3.75 & 2.99 & 4.09 & 3.34 & 4.53 & 3.98 & 4.47 & 4.26 \\
    \hline
90\% [px] & 7.39 & 6.01 & 7.52 & 6.50 & 6.98 & 5.28 & 7.42 & 5.80 & 7.04 & 6.08 & 7.75 & 7.12 \\
    \hline
99\% [px] & 11.19 & 8.88 & 11.33 & 9.58 & 9.67 & 8.08 & 9.60 & 8.27 & 9.97 & 8.10 & 10.81 & 10.05 \\
    \hline
99.9\% [px] & 11.85 & 9.72 & 12.23 & 12.07 & 11.15 & 9.63 & 11.07 & 10.04 & 10.83 & 8.95 & 12.89 & 13.27 \\
    \hline
    \hline
Mean [m]& 0.27 & 0.20 & 0.27 & 0.23 & 0.20 & 0.16 & 0.21 & 0.17 & 0.16 & 0.14 & 0.16 & 0.15\\
\hline     
\end{tabular}
\caption{\small{Accumulated frequency of the error: Depicted for all three altitudes, both training weights and corrections.}}
\label{tb:errorInPixel}
\end{table*}
\begin{figure}
\centering
\renewcommand{\arraystretch}{1.2}
\setlength{\tabcolsep}{2pt}
\noindent
\resizebox{0.5\textwidth}{!}{
\begin{tabular}{cc}
		\hline
		\large{Specialized training weights} & \large{General Training weights}\\
		\hline
		\noalign{\vskip 4mm}    
		\input{errorcurve_50m.tex} & \input{errorcurve_50m_no_A4.tex} \\
		\input{errorcurve_75m.tex} & \input{errorcurve_75m_no_A4.tex} \\
		\input{errorcurve_100m.tex} & \input{errorcurve_100m_no_A4.tex} \\
	\multicolumn{2}{c}{\ref{pl:regshift} \LARGE{reg+shift} \ref{pl:reg} \LARGE{reg} \ref{pl:raw} \LARGE{raw}}\\	
\end{tabular}
}
\caption{\small{Cumulative frequency diagrams: The left column depicts the results for the specialized training weights, the right column for the general training weights. The error above each plot is depicted in \textcolor{paperBlue}{[\unit{}{cm}]}, and in [\unit{}{px}] below each plot.}}
\label{fig:pos_est_fig}
\vspace{-0.4cm}
\end{figure}
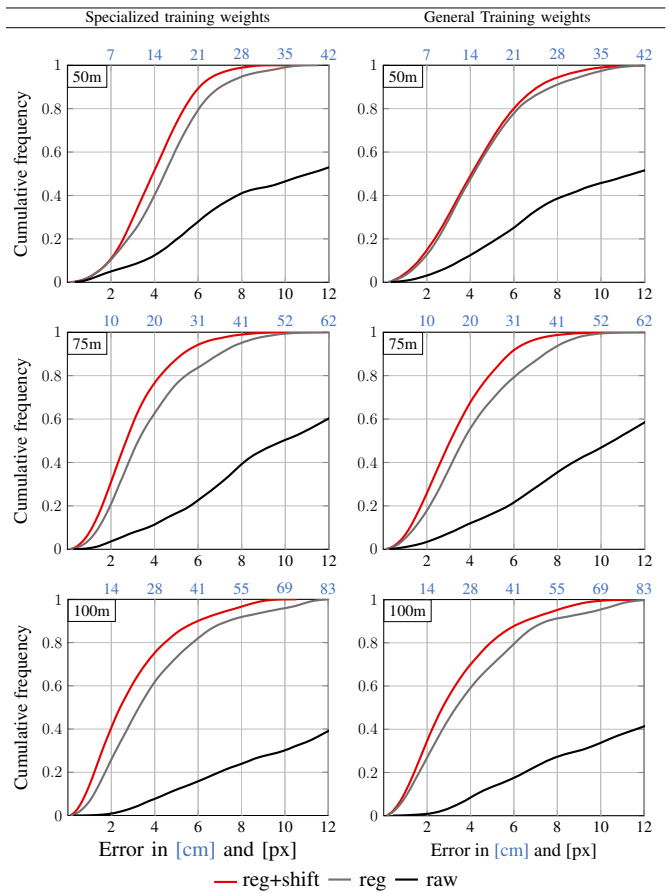
This section is concluded with the experimental results.

Figure \ref{fig:pos_est_fig} depicts the graph for each flight altitude, both training weights and the three main processing steps, where \ref{pl:raw} depicts results for non registered images, \ref{pl:reg} for registered images, and \ref{pl:regshift} for registered images with corrected relief displacement.

Image registration is the key to obtain reasonable results. The correction of the relief displacement improves the results by \unit[0.8]{px} on a weighted average\footnote{weighted by the number of frames per height, see Section \ref{architecture_section}}. Note, that the impact of the displacement is dependent on the distance $R$. Hence, data sets recording vehicles at the image border benefit more.

As mentioned before, the relief displacement is reduced with higher hovering altitudes. Remembering the issue regarding which vehicle part is actually corresponding to the outer pixel detected (Section \ref{architecture_section}), explains the best results in pixel measure for an altitude of \unit[100]{m}. However, expressing the error in the metric units, the error is lowest at \unit[50]{m}.

The experiments with the general training weights, which are based on images recorded on public roads, perform at weighted average only \unit[0.3]{px} worse. This underlines the suitability of the framework for applications on public roads. Table \ref{tb:errorInPixel} depicts a detailed comparison.

The mean error is \unit[20]{cm} and \unit[14]{cm} for a flight altitude of \unit[100]{m} and \unit[50]{m}, respectively. Regardless of the training weights, \unit[90]{\%} of all frames have an error of \unit[7]{px} or less.

Summarizing this section and the experimental results leads to the following conclusions:
\begin{inparaenum}[1)]
	\item A robust image registration is crucial for a good performance. 
	Omitting the effect of the relief displacement yields larger errors, when hovering above the region of interest is not feasable and objects are detected throughout the complete image frame.
	\item Considering the pixelwise results, similar performance can be observed for all three altitudes, which proofs that data can be obtained from different flight heights by a single Mask-RCNN network. This advantage can also be helpful for different object sizes.
	\item The best results in metric values are retrieved at lower altitudes. Alternatively, in order to capture a larger surface area, one can record at higher altitudes, increase the resolution and crop the image if necessary.
	\item Regarding real-world applications, the vehicle can at least be associated to its lane as exemplified in Figure \ref{fig:example_trajectory}. Note, that to some extend, the error values reported in Figure \ref{fig:pos_est_fig} and Table \ref{tb:errorInPixel} can be assigned to the synchronisation and mapping uncertainty, which is only of concern when benchmarking two data sources. 
	\item The vehicle to lane association also holds for the general training set, so that one can expect similar results for public roads, given an approriate training data set.
\end{inparaenum}	
\section{CONCLUSIONS}
\label{conclusions_section}
Vehicle detection by means of UAV imagery is an attractive option to generate data sets with relatively low effort.
This paper describes an approach based on deep neural network object detection and automated image registration techniques with state-of-the-art algorithms. A procedure to reduce the impact of relief displacement originated by the perspective projection of vertical images is described as well. Additionally, an overview of the potential sources of errors, and how to minimize their impact, is given.

The estimated vehicle position is compared to a reference system. Recording the data on a test track with consistent conditions ensures meaningful results. It is shown, that without applying any time-smoothing techniques, the position can be estimated in a reliable manner. The mean error is \unit[20]{cm} and \unit[14]{cm} for a flight altitude of \unit[100]{m} and \unit[50]{m}, respectively. Furthermore, \unit[90]{\%} of the 53\,855 independently evaluated frames have an error of \unit[7]{px} or less. To highlight the generalization capabilities, the experiments were analysed for two training data sets. One is a specialized data set, while the second being recorded on public roads. Both sets perform at a similar level. This allows the framework to be used for a wide range of applications. Interested readers are referred to the repository \cite{KruberGithub.2020}, where the code, label data and exmaple videos are made publicy available. 



\section*{ACKNOWLEDGMENT}
The authors acknowledge the financial support by the Federal Ministry of Education and Research of
Germany (BMBF) in the framework of FH-Impuls (project number 03FH7I02IA). The authors thank the AUDI AG department for Testing Total Vehicle for supporting this work.
\newpage 
\bibliographystyle{IEEEtran}
\bibliography{ref}
\end{document}

%% file: defs.tex
\definecolor{carBlue}{rgb}{0.2059,0.69412,0.83922}%
\definecolor{carRed}{rgb}{0.99608,0.29,0.23922}%
\definecolor{carGreen}{rgb}{0.63137,0.78431,0.25098}%
\definecolor{paperBlue}{rgb}{0.2656,0.4453,0.7656}%
\definecolor{paperGreen}{rgb}{0.63137,0.78431,0.25098}%
\definecolor{pathGreen}{rgb}{0.63137,0.78431,0.25098}%
\definecolor{pathRed}{rgb}{0.99608,0.44118,0.23922}%
\definecolor{paperRed}{rgb}{0.9,0.0,0.0}%
\definecolor{paperOrange}{rgb}{0.96078,0.89804,0.14902}%
\definecolor{paperGray}{rgb}{0.461,0.441,0.441}%

\newlength{\carLength}
\newlength{\freeLength}
\newlength{\fwidth}
\def\definemappedcolor#1{%
    \pgfmathparse{#1*1000}
    \pgfplotscolormapdefinemappedcolor{\pgfmathresult}%
}%

\pgfplotsset{
    colormap={accBrColor}{
        color=(carGreen)
        color=(carRed)
    },
    colormap={brAccColor}{
    	color=(carRed)
        color=(carGreen)
    },
    colormap={Acc1Color}{
    	color=(carGreen!50)
        color=(carGreen!50)
    },
    colormap={Acc2Color}{
    	color=(carGreen!70)
        color=(carGreen!70)
    },
    colormap={Acc3Color}{
    	color=(carGreen)
        color=(carGreen)
    },
    colormap={Br1Color}{
    	color=(carRed!50)
        color=(carRed!50)
    },
    colormap={Br2Color}{
    	color=(carRed!70)
        color=(carRed!70)
    },
    colormap={Br3Color}{
    	color=(carRed)
        color=(carRed)
    },
    colormap={Norm1Color}{
    	color=(carBlue!50)
        color=(carBlue!50)
    },
    colormap={Norm2Color}{
    	color=(carBlue!70)
        color=(carBlue!70)
    },
    colormap={Norm3Color}{
    	color=(carBlue)
        color=(carBlue)
    },
}

\tikzset{
  set arrow inside/.code={\pgfqkeys{/tikz/arrow inside}{#1}},
  set arrow inside={end/.initial=>, opt/.initial=},
  /pgf/decoration/Mark/.style={
    mark/.expanded=at position #1 with
    {
      \noexpand\definemappedcolor{#1}%
      \noexpand\arrow[\pgfkeysvalueof{/tikz/arrow inside/opt}]{\pgfkeysvalueof{/tikz/arrow inside/end}}
    }
  },
  arrow inside/.style 2 args={
    set arrow inside={#1},
    postaction={
      decorate,decoration={
        markings,Mark/.list={#2}
      }
    }
  },
}

\DeclareMathAlphabet{\mathsfit}{\encodingdefault}{\sfdefault}{m}{sl}
\SetMathAlphabet{\mathsfit}{bold}{\encodingdefault}{\sfdefault}{bx}{sl}

%% file: ToolChainComplete.tex
\begin{tikzpicture}
\def\ownSpace{4ex}
\def\ownHeight{6ex}
\def\ownWidth{23ex}

\node(Generation)[fill=gray!2,draw=paperRed, ultra thick]{
\begin{minipage}[t][\ownHeight][t]{\ownWidth}\vspace{0.5ex}\centering
\textbf{Data generation}\\[0.5ex] Video and DGPS
\end{minipage}};

\node(PreProcess)at($(Generation.east)+(\ownSpace,0)$)[anchor=west,fill=gray!2,draw=paperRed, ultra thick]{
\begin{minipage}[t][\ownHeight][t]{\ownWidth}\vspace{0.5ex}\centering
\textbf{Pre-Processing}\\[0.5ex] Image registration
\end{minipage}};

\node(Detection)at($(PreProcess.east)+(\ownSpace,0)$)[anchor=west,fill=gray!2,draw=paperRed, ultra thick]{
\begin{minipage}[t][\ownHeight][t]{\ownWidth}\vspace{0.5ex}\centering
\textbf{Object Detection}\\[0.5ex] Rotated Bounding Box
\end{minipage}};

\node(PostProcess)at($(Detection.east)+(\ownSpace,0)$)[anchor=west,draw=paperRed, ultra thick]{
\begin{minipage}[t][\ownHeight][t]{\ownWidth}\vspace{0.5ex}\centering
\textbf{Post-Processing}\\[0.5ex] Mapping, Relief D.
\end{minipage}};

\draw[-stealth,very thick] (Generation.east) -- (PreProcess.west);

\draw[-stealth,very thick] (PreProcess.east) -- (Detection.west);

\draw[-stealth,very thick] (Detection.east) -- (PostProcess.west);

\end{tikzpicture}

%% file: scale_and_theta_registration.tex
\begin{tikzpicture}
\footnotesize
\begin{axis}[
axis y line*=left,
grid=major,
width=0.98\columnwidth,
xlabel=$\text{time [s]}$,
ylabel=$\text{scaling factor}$,
xmin=0,
xmax=3000,
xticklabels={0,10,20,30,40,50,60},
xtick={0,500,1000,1500,2000,2500,3000},
xlabel near ticks,
ylabel near ticks,
  y tick label style={
    /pgf/number format/.cd,
    fixed,
    fixed zerofill,
    precision=3
  },
]

\addplot[color=paperRed] table [x= frame,y=scale_reg]{scale_registration_DJI0010.dat};\label{pl:scale_reg}

\end{axis}

\begin{axis}[
hide x axis,
axis y line*=right,
width=0.98\columnwidth,
ylabel=$\text{rotation [deg]}$,
ymin=0, 
xmin=0,
xmax=3000,
legend pos=south east,
xlabel near ticks,
ylabel near ticks
]

\addplot[color=black]  table [x=frame,y=theta_reg]{theta_registration_DJI0010.dat};\label{pl:theta_reg}

    \addlegendimage{/pgfplots/refstyle=pl:scale_reg}\addlegendentry{rotation}
    \addlegendimage{/pgfplots/refstyle=pl:theta_reg}\addlegendentry{scaling}

\end{axis}

\end{tikzpicture}

%% file: errorcurve_50m.tex
\begin{tikzpicture}
\large
\begin{axis}[
grid=major,
width=0.98\columnwidth,
ylabel=\Large{Cumulative frequency},
ymin=0, 
ymax=1,
xmin=0,
xmax=12,
xticklabels={2,4,6,8,10,12},
xtick={2,4,6,8,10,12},
xtick align=inside,
              extra x ticks={2,4,6,8,10,12},
      extra x tick labels={7,14,21,28,35,42},
      every extra x tick/.style={major tick length=0pt,
        xtick align=inside,ticklabel pos=top, color = paperBlue},
ylabel near ticks,
every axis plot/.append style={ultra thick},
title= \unit{50}{m},
every axis title/.style={below right,at={(0,1)},draw=black,fill=white}
]

\addplot[color=paperRed] table [x=error,y=percent]{plot_error_50m_reg_shift.dat};\label{pl:regshift}
\addplot[color=paperGray]  table [x=error,y=percent]{plot_error_50m_reg.dat};\label{pl:reg}
\addplot[color=black] table [x=error,y=percent]{plot_error_50m.dat};\label{pl:raw}
\end{axis}
\end{tikzpicture}

%% file: errorcurve_50m_no_A4.tex
\begin{tikzpicture}
\large
\begin{axis}[
grid=major,
width=0.98\columnwidth,
ymin=0, 
ymax=1,
xmin=0,
xmax=12,
xticklabels={2,4,6,8,10,12},
xtick={2,4,6,8,10,12},
xtick align=inside,
              extra x ticks={2,4,6,8,10,12},
      extra x tick labels={7,14,21,28,35,42},
      every extra x tick/.style={major tick length=0pt,
        xtick align=inside,ticklabel pos=top, color = paperBlue},
every axis plot/.append style={ultra thick},
title= \unit{50}{m},
every axis title/.style={below right,at={(0,1)},draw=black,fill=white}
]

\addplot[color=paperRed] table [x=error,y=percent]{plot_error_50m_reg_shift_no_A4.dat};\label{pl:regshift}
\addplot[color=paperGray]  table [x=error,y=percent]{plot_error_50m_reg_no_A4.dat};\label{pl:reg}
\addplot[color=black] table [x=error,y=percent]{plot_error_50m_no_A4.dat};\label{pl:raw}
\end{axis}
\end{tikzpicture}

%% file: errorcurve_75m.tex
\begin{tikzpicture}
\large
\begin{axis}[
grid=major,
width=0.98\columnwidth,
ylabel=\Large{Cumulative frequency},
ymin=0, 
ymax=1,
xmin=0,
xmax=12,
xticklabels={2,4,6,8,10,12},
xtick={2,4,6,8,10,12},
xtick align=inside,
              extra x ticks={2,4,6,8,10,12},
      extra x tick labels={10,20,31,41,52,62},
      every extra x tick/.style={major tick length=0pt,
        xtick align=inside,ticklabel pos=top, color = paperBlue},
ylabel near ticks,
every axis plot/.append style={ultra thick},
title= \unit{75}{m},
every axis title/.style={below right,at={(0,1)},draw=black,fill=white}
]

\addplot[color=paperRed] table [x=error,y=percent]{plot_error_75m_reg_shift.dat};\label{pl:regshift}
\addplot[color=paperGray]  table [x=error,y=percent]{plot_error_75m_reg.dat};\label{pl:reg}
\addplot[color=black] table [x=error,y=percent]{plot_error_75m.dat};\label{pl:raw}
\end{axis}
\end{tikzpicture}

%% file: errorcurve_75m_no_A4.tex
\begin{tikzpicture}
\large
\begin{axis}[
grid=major,
width=0.98\columnwidth,
ymin=0, 
ymax=1,
xmin=0,
xmax=12,
xticklabels={2,4,6,8,10,12},
xtick={2,4,6,8,10,12},
xtick align=inside,
              extra x ticks={2,4,6,8,10,12},
      extra x tick labels={10,20,31,41,52,62},
      every extra x tick/.style={major tick length=0pt,
        xtick align=inside,ticklabel pos=top, color = paperBlue},
every axis plot/.append style={ultra thick},
title= \unit{75}{m},
every axis title/.style={below right,at={(0,1)},draw=black,fill=white}
]

\addplot[color=paperRed] table [x=error,y=percent]{plot_error_75m_reg_shift_no_A4.dat};\label{pl:regshift}
\addplot[color=paperGray]  table [x=error,y=percent]{plot_error_75m_reg_no_A4.dat};\label{pl:reg}
\addplot[color=black] table [x=error,y=percent]{plot_error_75m_no_A4.dat};\label{pl:raw}
\end{axis}
\end{tikzpicture}

%% file: errorcurve_100m.tex
\begin{tikzpicture}
\large
\begin{axis}[
grid=major,
xlabel=$\text{\LARGE{Error in \textcolor{paperBlue}{[\unit{}{cm}]} and [\unit{}{px}]}}$,
ylabel=\Large{Cumulative frequency},
ymin=0, 
ymax=1,
xmin=0,
xmax=12,
xticklabels={2,4,6,8,10,12},
xtick={2,4,6,8,10,12},
xtick align=inside,
              extra x ticks={2,4,6,8,10,12},
      extra x tick labels={14,28,41,55,69,83},
      every extra x tick/.style={major tick length=0pt,
        xtick align=inside,ticklabel pos=top, color = paperBlue},
ylabel near ticks,
every axis plot/.append style={ultra thick},
title= \unit{100}{m},
every axis title/.style={below right,at={(0,1)},draw=black,fill=white}
]

\addplot[color=paperRed] table [x=error,y=percent]{plot_error_100m_reg_shift.dat};\label{pl:regshift}
\addplot[color=paperGray]  table [x=error,y=percent]{plot_error_100m_reg.dat};\label{pl:reg}
\addplot[color=black] table [x=error,y=percent]{plot_error_100m.dat};\label{pl:raw}
\end{axis}
\end{tikzpicture}

%% file: errorcurve_100m_no_A4.tex
\begin{tikzpicture}
\large
\begin{axis}[
grid=major,
xlabel=$\text{\Large{Error in \textcolor{paperBlue}{[\unit{}{cm}]} and [\unit{}{px}]}}$,
ymin=0, 
ymax=1,
xmin=0,
xmax=12,
xlabel near ticks,
xticklabels={2,4,6,8,10,12},
xtick={2,4,6,8,10,12},
xtick align=inside,
              extra x ticks={2,4,6,8,10,12},
      extra x tick labels={14,28,41,55,69,83},
      every extra x tick/.style={major tick length=0pt,
        xtick align=inside,ticklabel pos=top, color = paperBlue},
every axis plot/.append style={ultra thick},
title= \unit{100}{m},
every axis title/.style={below right,at={(0,1)},draw=black,fill=white}
]

\addplot[color=paperRed] table [x=error,y=percent]{plot_error_100m_reg_shift_no_A4.dat};\label{pl:regshift}
\addplot[color=paperGray]  table [x=error,y=percent]{plot_error_100m_reg_no_A4.dat};\label{pl:reg}
\addplot[color=black] table [x=error,y=percent]{plot_error_100m_no_A4.dat};\label{pl:raw}
\end{axis}
\end{tikzpicture}